\documentclass{article}

\usepackage{times,fullpage}

\usepackage[utf8]{inputenc} % allow utf-8 input
\usepackage[T1]{fontenc}    % use 8-bit T1 fonts
\usepackage[pagebackref=true,breaklinks=true,letterpaper=true,colorlinks,bookmarks=false]{hyperref}

\usepackage{url}            % simple URL typesetting
\usepackage{booktabs}       % professional-quality tables
\usepackage{amsfonts}       % blackboard math symbols
\usepackage{nicefrac}       % compact symbols for 1/2, etc.
\usepackage{microtype}      % microtypography
\usepackage{multicol}
\usepackage[title]{appendix}

\usepackage{comment}
\usepackage{float}
\usepackage{graphics}
\usepackage{graphicx}
\usepackage[font=small,labelfont=bf]{caption}
\usepackage{subcaption}
\usepackage{enumerate}
\usepackage{amsmath}
\usepackage{amssymb}
\usepackage{pdfpages}
\usepackage{sidecap}

\usepackage{verbatim}

\usepackage[linesnumbered,ruled,vlined,commentsnumbered]{algorithm2e}
\makeatletter
\g@addto@macro{\@algocf@init}{
    \SetFuncSty{textsc}
    \SetKwInOut{Input}{Input}
    \SetKwInOut{Output}{Return}
    \SetKwProg{Procedure}{Procedure}{}{}
    \SetKw{Continue}{continue}
    \SetKw{Break}{break}
}
\makeatother

\usepackage[disable]{todonotes}
\usepackage{marginnote}

\title{PODDP: Partially Observable Differential Dynamic Programming for Latent Belief Space Planning}

\author{
  Dicong Qiu, Yibiao Zhao, Chris L. Baker \\
  isee.ai \\
  Cambridge, MA 02139 \\
  \texttt{\{dq,yz,chrisbaker\}@isee.ai} \\
}

\begin{document}

\maketitle

\begin{abstract}
Autonomous agents are limited in their ability to observe the world state. Partially observable Markov decision processes (POMDPs) formally model the problem of planning under world state uncertainty, but POMDPs with continuous actions and nonlinear dynamics suitable for robotics applications are challenging to solve.
In this paper, we present an efficient differential dynamic programming (DDP) algorithm for belief space planning in POMDPs with uncertainty over a discrete latent state, and continuous states, actions, observations, and nonlinear dynamics. This representation allows planning of dynamic trajectories which are sensitive to structured uncertainty over discrete latent world states.
We develop dynamic programming techniques to optimize a contingency plan over a tree of possible observations and belief space trajectories, and also derive a hierarchical version of the algorithm.
Our method is applicable to problems with uncertainty over the cost or reward function (e.g., the configuration of goals or obstacles), uncertainty over the dynamics (e.g., the dynamical mode of a hybrid system), and uncertainty about interactions, where other agents' behavior is conditioned on latent intentions. Benchmarks show that our algorithm outperforms popular heuristic approaches to planning under uncertainty, and results from an autonomous lane changing task demonstrate that our algorithm can synthesize robust interactive trajectories.
\end{abstract}

\section{Introduction}\label{sec:Introduction}

Planning under uncertainty resulting from a limited ability to observe the world state is a critical capacity for autonomous agents.
Noisy actuators, imperfect sensors, and perceptual limitations such as occlusion contribute to the uncertainty that agents face when deciding to act. Even with perfect sensors and perception, latent states of the world can remain opaque, such as whether a cookie jar is empty, or whether another driver intends to yield.
Planning under this uncertainty requires balancing the cost of exploratory actions with the potential benefit of the information gained.
However, the problem of planning under partial observability, which can be formalized as a partially observable Markov decision process (POMDPs), is generally intractable~\cite{papadimitriou1987}.

Trajectory optimization techniques
have proven very effective for robotics applications when embedded within nonlinear model-predictive control architectures~\cite{tassa2012}, but typically require the state to be fully observable (or separately estimated).
Extensions of trajectory optimization techniques to belief space planning allow partial observability to be captured within continuous motion planning algorithms suitable for robotics applications~\cite{vandenberg2012,patil2015}. However, these approaches primarily consider unimodal Gaussian uncertainty, which limits their utility for problems with multimodal structure. Real world uncertainty is multimodal: e.g., about the presence or location of a goal object, about the discrete mode of a system, or about the type of person with whom one is interacting. This multimodal structure can be represented within general POMDPs, which can be solved by general-purpose solvers~\cite{kurniawati2008,ye2017}. However, optimizing continuous actions for motion planning is challenging for state of the art POMDP solvers.

In this paper, we propose a trajectory optimization approach for solving nonlinear, continuous state, action, and observation POMDPs with non-Gaussian beliefs over discrete latent variables. Our approach, called Partially Observable Differential Dynamic Programming (PODDP), builds and optimizes a contingency plan over a tree of possible observations and trajectories in the belief space.
We derive techniques for dynamic programming over the trajectory tree, which involve propagating an approximate value function through the belief state dynamics defined by observations and Bayesian belief updating. Lastly, we describe a hierarchical dynamic programming decomposition of the problem, which is practical for robotics settings, where receding horizon planning is applicable, and where the control frequency is typically higher than the observation frequency.

The PODDP representation and method allow us to model and optimize trajectories for several important classes of nonlinear, continuous planning problems with uncertainty over discrete latent states:
(1) Tasks where the cost function depends on an uncertain latent state, e.g., where an agent must approach or avoid goals or obstacles which may be in a finite number of locations.
(2) Tasks where the dynamics are conditioned on the uncertain latent mode of the (hybrid) system, e.g., contact mode, component status, or environmental condition (e.g., smooth vs. rough terrain).
(3) Interactive tasks where other agents' trajectories impose dynamic costs and are influenced by their latent intentions, e.g., autonomous driving systems must plan under uncertainty about other vehicles' interactive trajectories, conditioned on their drivers' situational awareness level, intention to cooperate, etc.

We demonstrate the efficacy of our method experimentally in instances of each of these problem classes. First, we benchmark our algorithm in two partially observable domains against two alternative heuristic approaches: 1) a ``fully observable'' heuristic under which DDP treats the maximum-likelihood state as the true world state, and 2) a cost-based heuristic which combines multiple latent state hypotheses inside the cost function, weighted by their probabilities.
Finally, we consider a challenging autonomous driving setting in which the model must plan interactive lane changing trajectories, and show that PODDP can plan and execute successful lane change trajectories by inferring whether another agent intends to yield.

\section{Related work}

In this section we first review related research on planning under partial observability, with a focus on methods for planning in continuous action spaces. We review techniques for Gaussian belief space planning, as well as general-purpose POMDP solvers which can plan continuous actions. Finally, we review research applying related techniques, primarily in applications related to autonomous driving.

\subsection{Gaussian belief space planning}

Trajectory optimization methods compute a sequence of continuous actions to minimize the expected cost of the resulting state sequence.
For the special case of problems with linear dynamics, quadratic costs, and additive Gaussian process and observation noise, optimal continuous feedback policies can be computed by the LQG algorithm using dynamic programming~\cite{bertsekas2017}. In the LQG setting, the separation principle~\cite{astrom1970} implies that the same feedback policy is optimal when applied to either the true state in the fully observable setting, or to the estimated state in the partially observable setting.
In the nonlinear setting, differential dynamic programming (DDP)~\cite{jacobson1970} and iterative LQG (iLQG)~\cite{todorov2005,tassa2012} extend the dynamic programming approach to compute locally optimal feedback policies for systems with smooth, nonlinear dynamics and non-quadratic costs.
Like LQG, these algorithms separate estimation and control, and thus they are unable to plan exploratory actions, because they do not explicitly model the effect of observations on the belief dynamics.

Methods for trajectory optimization in Gaussian belief space~\cite{platt2010belief,vandenberg2012,erez2010,patil2015} model the belief dynamics induced by Gaussian process and observation noise by augmenting the state to include the estimated mean and covariance, and propagating the belief state with a Kalman filter. These methods use two main ways to approximate the belief dynamics. The first approximation assumes that the observations from each state take their maximum-likelihood values~\cite{platt2010belief,erez2010,patil2015}, which makes the belief space dynamics deterministic. This heuristic reduces to the LQG algorithm in the linear-quadratic-Gaussian setting, and provides a lower bound on the uncertainty of future belief states in the nonlinear setting.
The second approximation propagates the belief state by
propagating the linearized Gaussian belief state dynamics within each step of dynamic programming, and~\cite{vandenberg2012} showed that this approach outperforms the maximum-likelihood observations heuristic.
Both approximation methods yield policies which explore the environment to gain information that is useful for optimizing future value.

Other techniques use LQG controllers to facilitate sampling-based Gaussian belief space motion planning. \cite{vandenberg2011} and~\cite{bry2011} use LQG and Kalman filtering to estimate the expected covariance, cost, and success probability of tracking trajectories computed by RRT in the presence of motion and sensor noise. The results of both methods demonstrate that they are able to plan information gathering, exploratory actions. The FIRM algorithm~\cite{agha2011} constructs an information-state roadmap, an extension of probabilistic roadmaps~\cite{kavraki1996} to problems with motion and sensing uncertainty, where nodes and edges lie in the belief space, and uses LQG controllers to break the curse of history and guarantee reachability of nodes in the roadmap.

There are several reinforcement learning techniques (RL) for solving continuous state-action-observation POMDPs. The model-based approach of~\cite{mcallister2017} extends the PILCO algorithm to continuous state-action Gaussian POMDPs, by simultaneously learning a Gaussian process model dynamical model and policy. \cite{igl18} apply model-free deep variational RL to jointly learn a dynamical model and a continuous policy, and apply this method to a Gaussian mountain hike problem.

Gaussian belief space planning assumes that all uncertainty can be represented in the form of unimodal Gaussian distributions over the state space. In contrast, our work here captures the multimodality of real-world uncertainty.
In a similar spirit to our work,~\cite{jain2017} propose an extension of the belief space planning method of~\cite{platt2010belief} to hybrid continuous dynamics, in which a (partially observable) discrete mode determines the continuous dynamics of the system. Several other techniques have been developed for this setting, including the Sequential Action Control approach of~\cite{nishimura2018sacbp}, and the point-based method of~\cite{brunskill2008} (which assumes discrete actions).

\subsection{General purpose POMDP solvers}

Point-based solvers~\cite{kurniawati2008} allow medium-sized POMDPs to be solved offline, and have been extended to continuous state spaces~\cite{porta2006}, but require discretization of the action and observation spaces. Monte-Carlo tree search (MCTS) is an online method which can scale to large domains~\cite{silver2010,ye2017}. MCTS can naturally handle continuous state spaces, but continuous observations and actions are more complicated. Fine discretization of the observation space is feasible because MCTS can scale to large observation spaces; however, MCTS scales exponentially in the number of actions, so only coarse discretization of actions is tractable~\cite{cai2018}. Several techniques have been proposed to handle continuous actions and observations. The approach of~\cite{sunberg2017} applies double progressive widening~\cite{couetoux2011} to actions and observations in MCTS, while~\cite{seiler2015} performs local search in the belief tree for optimal continuous actions.

\subsection{Related applications}

The active SLAM problem is similar in spirit to the problem classes we consider involving uncertainty about the spatial structure of the environment (see Experiments 1 and 2). The large literature in this area~\cite{cadena2016} is outside the scope of this review. \cite{indelman2015} present a belief space planning approach to the active SLAM problem, and augment Gaussian dynamics and observations with binary random variables representing whether a measurement is taken. \cite{patil2015} are able to scale their belief space planner to 50 landmarks in an active SLAM experiment.

There large literature on interactive and intention-aware planning approaches for mobile robots is highly relevant to our work. Although planning for interaction using a world model that includes (a distribution over) other agents' predicted trajectories as part of the dynamics -- treating other agents as ``part of the environment'' -- is a popular technique~\cite{bai2015,sadigh2016,hardy2013contingency, galceran2017,fisac2019}, planning over uncertain predictions of other agents' trajectories can lead to the freezing robot problem~\cite{trautman2010}, while treating other agents' trajectories as deterministically predictable can lead to overly aggressive behavior~\cite{sadigh2016,fisac2019}. A variety of solutions have been proposed.
Intention-aware POMDP formulations of interaction are conceptually similar to the approach we take: they solve a POMDP with a belief state defined by hidden intent variables which influence the predicted trajectory for each agent~\cite{bai2015,sunberg2017,galceran2017}.  \cite{schwarting2019} considers interactive behavior that is governed by a continuous latent variable which determines other drivers' ``social value orientation'' (SVO. Their algorithm plans a game-theoretic equilibrium, given SVO estimates, for all vehicles simultaneously using multi-agent trajectory optimization techniques. We take inspiration from the approach of trajectory-optimization through differentiable models of other agents~\cite{sadigh2016}; our agent models are greatly simplified, but they are enhanced to allow intent uncertainty. The algorithm we propose is structurally similar to the contingency-planning approach of~\cite{hardy2013contingency}; however, our method operates over observation horizons longer than one step, and takes a DDP-based approach to trajectory optimization.

\section{Problem formulation}

We consider finite-horizon trajectory planning and control problems, in environments with hybrid continuous and discrete state, and continuous actions and observations. We factorize the state space $\mathcal{S} = \mathcal{X} \times \mathcal{Z}$ into a continuous state space $\mathcal{X}$, and a discrete state space $\mathcal{Z}$. The discrete state represents the latent mode of the hybrid system. The control space $\mathcal{U}$, and observation space $\mathcal{O}$ are continuous.
In this paper, for simplicity we assume the continuous state is fully observable, and only the discrete state is partially observable; our formulation is thus a mixed-observability MDP model~\cite{ong2009}, which yields more compact representations of the belief space and dynamics. However, we note that our model can be extended to allow partially observable continuous state using well-known techniques~\cite{van2017motion}.
We also assume for simplicity that the value of the hidden state is constant over the planning horizon. This is sensible in our setting because local methods for trajectory optimization are best suited to problems with relatively short planning horizons (e.g. model-predictive control), over which the hidden state can be assumed to be constant. Technically, this makes our formulation a POMDP-lite~\cite{chen2016}; however, the extension to stochastic dynamics over the hidden state is straightforward.

The system dynamics are defined by the conditional distribution over the next state $x_{t+1}$, $p(x_{t+1}| x_t,u_t,z)$, which depends on the current state $\langle x_t, z \rangle \in \mathcal{S}$, and control $u_t\in\mathcal{U}$.
The observation distribution $p(o_t|x_t, z)$ is also conditioned on the current state.

The current belief about the hidden state depends on the history of observed states, controls, and observations. We use recursive Bayesian filtering to update the current belief, based on the latest observation, which includes both $o_t$ and $x_t$, following the mixed-observability assumption:
\begin{equation}
\begin{split}
    b_{t+1}(z)
    &\doteq P(z|o_t,x_t,u_{t-1},\ldots,o_1,x_1,u_0,x_0,b_0) \\
    &= P(z|o_t,x_t,u_{t-1},x_{t-1},b_t) \\
    &= \eta \cdot p(o_t|x_t,z) p(x_t|x_{t-1},u_{t-1},z) b_t(z) \\
    &= h(o_t,x_t,u_{t-1},x_{t-1},b_t),
\end{split}
\label{eqn:BeliefUpdate}
\end{equation}
where $\eta$ is a normalizing constant, and where we define $h()$ to denote the deterministic belief update function mapping $b_t$ to $b_{t+1}$.

The running loss function $l(x_t,u_t,z)$ represents the loss incurred by the control $u_t$ from the current state, and the final loss function $l_f(x_T,z)$ represents the loss incurred within a state at the end of the planning horizon; both functions are assumed to be twice-differentiable. To minimize costs, controls are conditioned on the current belief state, defined as the pair $\langle x_t, b_t \rangle$, because the current POMDP state $\langle x_t, z \rangle$ is not fully observable. We define the expected finite-horizon cost of a policy $\pi$ mapping belief states to controls recursively:
\begin{equation}
\begin{split}
    V^{\pi}(x_t,b_t) &=
    \mathbb{E}_{z\sim b_t} \left[l(x_t,\pi(x_t,b_t),z)
    + \mathbb{E}_{o_t,x_{t+1}} [ V^{\pi}(x_{t+1},h(o_t,x_t,u_{t-1},x_{t-1},b_t))] \right] \\
    V^{\pi}(x_T,b_T) &=
    \mathbb{E}_{z\sim b_T} [l_f(x_T,z)],
\end{split}
\label{eqn:ValueFunction}
\end{equation}
where in the second expectation $o_t \sim p(o_t|x_t,z)$ and $x_{t+1} \sim p(x_{t+1}|x_t,u_t,z)$, and where the value at the planning horizon $T$ is the expected final cost.
The optimal policy is defined as:
\begin{equation}
    \pi^*(x,b) = {\arg\max}_{\pi} V^{\pi}(x,b).
\end{equation}
In the next section we describe our PODDP method for computing $V$ and $\pi^*$.

\section{Partially observable differential dynamic programming}

Standard DDP optimizes a trajectory by alternating a forward pass which rolls out the dynamics and costs using a control sequence, and a backward pass which takes a local second-order approximation to the value function, and updates the control sequence to optimize the approximate value function. This process repeats until a locally optimal trajectory is found. Dynamic programming algorithms for Gaussian belief space planning~\cite{platt2010belief,van2017motion} proceed similarly to standard DDP -- they roll out a trajectory in the forward pass by propagating a belief state defined by the mean and variance of the state, and optimize an approximate value function around the belief state trajectory in the backward pass.

\begin{SCfigure}[][htb]
    \centering
    \includegraphics[width=0.42\linewidth]{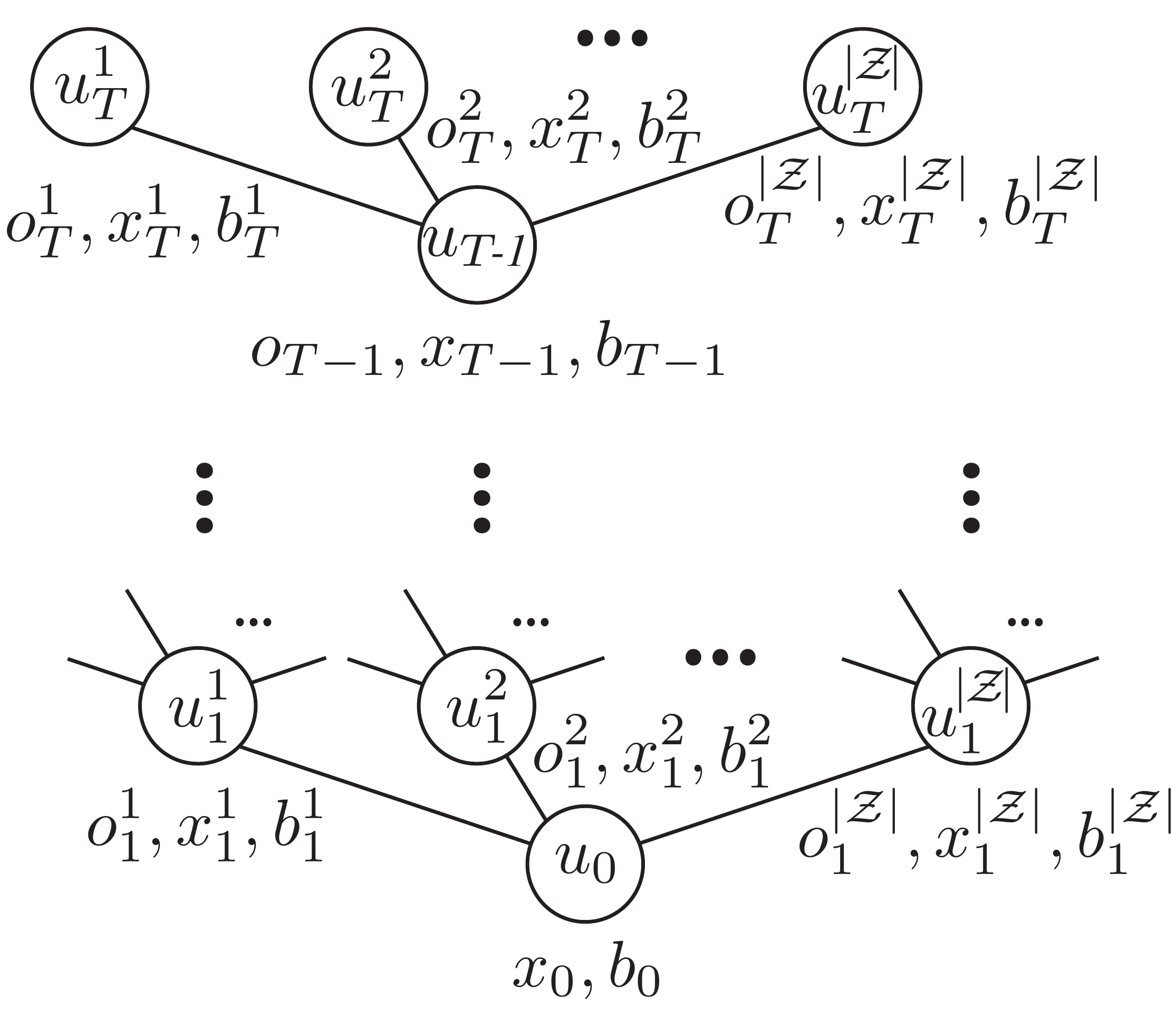}
    \caption{PODDP trajectory tree. Starting from belief state $\langle x_0, b_0 \rangle$, tree construction rolls out control $u_0$ for each possible latent state value $z \in \mathcal{Z}$, assuming next state $x_1^z$ and observation $o_1^z$ take their maximum likelihood values, and $b_1^z$ is given by Bayesian belief updating. Tree construction proceeds recursively from each $x_1^z, b_1^z$ until the finite horizon is reached. Note that in the last layer we have suppressed the superscript labels for $o_{T-1}, x_{T-1}, b_{T-1}$ for clarity -- the notation is cumbersome, and should record the complete history of latent state values used to generate the state and observation sequence preceding the node.}
    \label{fig:PODDP_policy_tree}
\end{SCfigure}

PODDP plans in belief space, but unlike Gaussian belief space planning, the marginal distribution over observations is not unimodal, and the belief-space trajectory cannot be approximated by propagating a single sequence of means and variances. In our setting, the discrete latent variable $z$ induces a multimodal distribution over observations, and a non-Gaussian belief state, which induces a(n infinitely branching) tree of observations, beliefs, and controls in the forward pass. The PODDP forward pass approximates this tree with a finite representation which we call a ``trajectory tree'', shown in Fig.~\ref{fig:PODDP_policy_tree}, analogous to the policy tree representation in the classic discrete POMDP setting~\cite{kaelbling1998}. The PODDP backward pass proceeds from the leaves of the tree, and propagates the value through observations and belief updates via dynamic programming.
The remainder of this section will derive the forward and backward passes of the PODDP algorithm, and then derive an efficient hierarchical decomposition of the trajectory tree.

\subsection{PODDP forward pass}
\label{sec:ForwardPass}

Given an initial belief state $\langle x_0, b_0 \rangle$, the PODDP forward pass constructs a trajectory tree which approximates the infinite space of possible control, state, observation, and belief sequences up to a finite horizon $T$. Each node in the tree is labeled by the control to be executed if that node is reached. From each node, we generate a finite set of branches corresponding to possible state transitions, observations, and belief updates given the control and belief state at that node. A control node is created following each branch, and tree expansion proceeds recursively until the finite horizon is reached. Fig.~\ref{fig:PODDP_policy_tree} sketches the trajectory tree structure, and Algorithm~\ref{alg:ForwardPass} defines the algorithm formally.

\begin{algorithm}[bht]
    \SetKwFunction{ForwardTree}{ForwardTree}
    \SetKwFunction{BeliefUpdate}{BeliefUpdate}
    \SetKwFunction{UpdateBelief}{UpdateBelief} % external

    $ U \leftarrow [\,] $ \tcp*{initialize control map indexed by histories}
    $ S(\textit{`Root'}) \leftarrow [x_0, b_0] $ \tcp*{initialize belief state map indexed by histories}

    $ \ForwardTree\left(\textit{`Root'}, U, S, U_{\text{nom}}, S_{\text{nom}}, \varnothing, \varnothing, \alpha, \mathcal{Z}, T,  1 \right) $ \tcp*{trajectory tree recursion}
    \Return $ U, S $ \tcp*{return updated trajectory tree}

    \Procedure{\ForwardTree{$ H, U, S, U_{\text{nom}}, S_{\text{nom}}, k, K, \alpha, \mathcal{Z}, T, d $}}{
        \uIf(\tcp*[f]{apply control updates}){$ k( H ), K( H ) \neq \varnothing $}{
            $ U( H ) \leftarrow U_{\text{nom}}( H ) + \alpha k( H ) + K( H ) ( S(H) - S_{\text{nom}}( H ) )$ \;
        }
        \Else{$ U(H) \leftarrow U_{\text{nom}}(H)$ \;
        }

        \For{$ z \in \mathcal{Z} $}{
            $ [x_H,b_H] \leftarrow S(H) $ \;
            $x' = \arg\max_x p(x| x_H, U(H), z)$ \tcp*{assume ML state transition}
            $o' = \arg\max_o p(o| x', z)$ \tcp*{assume ML observation}
            $b' = \BeliefUpdate(o', x', U(H), x_H, b_H)$ \;
            $S([H,z]) \leftarrow [x', b']$ \tcp*{append new belief state to history}
            \If{$ d < T $}{
                $\ForwardTree\left( [H, z], U, S, U_{\text{nom}}, X_{\text{nom}}, k, K, \alpha, \mathcal{Z}, T, d + 1 \right) $ \tcp*{recurse}
            }
        }
    }
    \caption{$ \textsc{PODDPForwardPass}\left( x_{0}, b_{0}, U_{\text{nom}}, S_{\text{nom}}, k, K, \alpha, \mathcal{Z}, T \right) $}
    \label{alg:ForwardPass}
\end{algorithm}

To approximate the infinite set of continuous observations that are possible from each node, we introduce a maximum-likelihood outcomes (MLO) heuristic~\cite{platt2010belief}: For each possible latent state value $z \in \mathcal{Z}$, we compute the maximum likelihood state transition and observation, and perform the belief update as defined in lines 12-14 of Algorithm~\ref{alg:ForwardPass}.
The MLO heuristic transforms the operation of sampling next states and observations into a deterministic function, which PODDP requires to be differentiable. To enable this, we assume that $p(x_t| x_{t-1}, u_{t-1}, z)$ and $p(o_t|x_t,z)$ are Gaussian distributions with additive noise; MLO corresponds to taking the mean of the distributions, and differentiation involves taking the derivative of the underlying process.

The forward pass is called on every iteration of PODDP. On the first iteration, the nominal controls $U_{\text{nom}}$ are initialized to a default value (constant in our examples in this paper; more complex schemes are possible), and nominal belief states $S_{\text{nom}}$ and the control updates $k$ and $K$ are set to null. At later iterations, $k$ and $K$, computed by the backward pass, specify modifications to the previous control $U_{\text{nom}}$, and provide linear feedback control gains to stabilize the trajectory around $S_{\text{nom}}$, respectively. The step size $\alpha$ is set by a line search procedure~\cite{tassa2012}. \todo{we actually are not using the improved line search from \cite{tassa2012}}

\subsection{PODDP backward pass}
\label{sec:BackwardPass}

DDP computes a second-order approximation to the value function around within a local region around a trajectory. Starting from the end of trajectory, DDP takes second-order approximations to the cost function and the dynamics, and uses dynamic programming to propagate the value function backward through time. During this process, locally optimal control modifications and linear feedback control gains are computed, which are used to update the trajectory during the next forward pass.

The PODDP backward pass operates over a trajectory tree, proceeding from the leaves, and propagating the value through the observations and belief updates that take place at each node.  Algorithm~\ref{alg:BackwardPass} defines this procedure, which traverses the trajectory tree in depth-first order, and propagates the necessary derivatives backward through the tree recursively. Next, we derive the remaining core function of Algorithm~\ref{alg:BackwardPass}, which performs the second order approximation to the value function, and returns the derivatives and control updates to be propagated backward through the trajectory tree.

\begin{algorithm}[htb]
    \SetKwFunction{BackwardTree}{BackwardTree}
    \SetKwFunction{OptimizeCtrl}{OptimizeControl} % external
    \SetKwFunction{Parent}{Parent} % operation
    \SetKwFunction{Child}{Child} % operation
    \SetKwFunction{History}{History} % field access operation

    $k, K \leftarrow [\,] $ \tcp*{initialize control update maps indexed by histories}
    $ \BackwardTree\left( \textit{`Root'}, k, K, U, S, \mathcal{Z}, T, 1 \right) $ \tcp*{compute control updates recursively}
    \Return $ k, K $ \tcp*{return control updates}

    \Procedure{\BackwardTree{$ H, k, K, u, S, \mathcal{Z}, T, d$}}{
        $\Delta \leftarrow [\,]$ \tcp*{initialize backward derivatives map}
        \For{$ z \in \mathcal{Z} $}{
            \uIf(\tcp*[f]{recursively compute backward derivatives and updates}){$ d < T $}{
                $ \Delta(z) \leftarrow \BackwardTree( [H,z], k, K, U, S, \mathcal{Z}, T, d ) $ \;
            }\Else(\tcp*[f]{set backward derivatives as empty}){
                $ \Delta(z) \leftarrow \varnothing $ \;
            }
        }

        $ k_H, K_h, \Delta_H \leftarrow \OptimizeCtrl( U( H ), S( H ), \Delta, \mathcal{Z} ) $ \;
        $ k( H ), K( H) \leftarrow k_H, K_H $ \tcp*{update control update maps}
        \Return $ \Delta_H $ \tcp*{return backward derivatives}
    }
    \caption{$ \textsc{PODDPBackwardPass}\left( U, S, \mathcal{Z}, T \right) $}
    \label{alg:BackwardPass}
\end{algorithm}

\subsubsection{Backward control updates and derivatives} \label{sec:QFunction}

Dynamic programming over the trajectory tree requires differentiation through the belief space dynamics at each observation and belief update. However, differentiating the raw belief state is problematic, because perturbations can push the belief off of the $|\mathcal{Z}|{-}1$-dimensional simplex.

To solve this issue, we reparameterize the belief state in terms of the unconstrained parameter $\beta \in \mathbb{R}^{|\mathcal{Z}|}$, such that:
\begin{equation}
b\left( z ; \beta \right) = \frac{ \exp\left( \beta\left( z \right) \right) }{ \sum_{z' \in \mathcal{Z}} \exp\left( \beta\left( z' \right) \right) },
\label{eqn:BeliefParam}
\end{equation}
and the belief space takes the form $\mathcal{S} = \mathcal{X} \times \mathbb{R}^{|\mathcal{Z}|}$. The reparameterized belief update naturally derives from Equation~\ref{eqn:BeliefUpdate}, such that $\beta_{t+1}(z) = \log(b_{t+1}(z))$.

To complete the derivation of the backward pass for PODDP, we define the state-action value function to operate over perturbations of $s$ and $u$:
\begin{eqnarray*}
Q\left( \delta s, \delta u \right)
    &=& \sum_{z \in \mathcal{Z}}
        b( z ; \beta{+}\delta\beta) \left[ l( x{+}\delta x, u{+}\delta u, z )
        + V\bigl( x{+}\delta x, h(o',x',u{+}\delta u,x{+}\delta x, b(\beta{+} \delta \beta)) \bigr) \right] \\
    &=& \sum_{z \in \mathcal{Z}} b_z(l_z + V(s_z'))
\label{eqn:Qfunction}
\end{eqnarray*}
where we have implicitly decomposed $\delta s$ into $\delta x$ and $\delta \beta$, and where we assume that $o'$ and $x'$ take on their maximum likelihood values. The second expression introduces variable abbreviations that we use below; also let $V_z' = V(s_z')$.

We take a second order approximation, $\tilde{Q}$ to the state-action value function by computing first- and second-derivatives with respect to $\delta s$ and $\delta u$. We present the first derivatives in the main text to show their structure; the Hessians are derived in Supplementary. We note that in this work, we employ the standard iLQR approach of discarding the Hessians of the dynamics~\cite{tassa2012}.
\begin{eqnarray}
Q_{s} &=& \sum_{z \in \mathcal{Z}} \left[
    \frac{ \partial b_{z} }{ \partial \delta s } \left( l_{z} + V_{z}' \right) +
    b_{z} \left(
        \frac{ \partial l_{z} }{ \partial \delta s } +
        \frac{ \partial s_{z}' }{ \partial \delta s }^{\mathsf{T}} \frac{ \partial V_{z}' }{ \partial s_{z}' }
    \right)
\right] \\
Q_{u} &=& \sum_{z \in \mathcal{Z}} \left[
    b_{z} \left(
        \frac{ \partial l_{z} }{ \partial \delta u } +
        \frac{ \partial s_{z}' }{ \partial \delta u }^{\mathsf{T}} \frac{ \partial V_{z}' }{ \partial s_{z}' }
    \right)
\right].
\end{eqnarray}
Although we differentiate the raw belief $b_z$, the reparameterization in Equation~\ref{eqn:BeliefParam} makes these derivatives well-behaved near the simplex boundary, where the derivatives take on small values for extremal beliefs, and small perturbations $\delta s$ do not violate the simplex constraint. The $\partial s_{z}' / \partial \delta s$ and $\partial s_{z}' / \partial \delta u$ terms involve differentiating through the dynamics, observation model, and belief update. The $\partial V_{z}'/\partial s_{z}'$ and $\partial V_{z}'/\partial s_{z}'$ terms are the backward derivatives propagated within the $\Delta$ argument in Algorithm~\ref{alg:BackwardPass}; we discuss how they are computed in Supplementary.

The optimal control modification $ \delta u^{*} $ for belief state perturbation $ \delta s $ is computed by minimizing the quadratic model $\tilde{Q}$:
\begin{equation}
\delta u^{*}\left( \delta s \right) = \arg\min_{u} \tilde{Q}\left( \delta s, \delta u \right) = k + K \delta s,
\label{eqn:ControlUpdate}
\end{equation}
where $ k = - Q_{uu}^{-1} Q_{u} $ is an open-loop modification to be applied in the forward pass, and $ K = - Q_{uu}^{-1} Q_{us} $ is a linear closed-loop feedback gain.

\subsection{Hierarchical PODDP}
\label{sec:HierarchicalPODDP}

Because each node in the trajectory tree has $|\mathcal{Z}|$ successor nodes, the tree has size $(|\mathcal{Z}|^T-1) / (|\mathcal{Z}|-1) = \mathcal{O}(|Z|^T)$. This exponential growth is manageable for short horizons ($T < 5)$, but for longer horizons typically required for robotics tasks, it is infeasible. However, branching on every timestep may be unnecessary for several reasons. First, many robotics systems have high control frequency, but much lower state estimation frequency, particularly for sensor fusion from multiple modalities (i.e., cameras, LIDAR, etc.) In this case, it makes sense to align the observation timesteps in the planner with those when observations are expected to occur in the system. Second, planning with a lower observation frequency than that of the actual system can yield trajectories which take observation contingencies into account, but are more conservative than those which observe at every timestep.

To derive the hierarchical PODDP algorithm, we follow the derivation above, but partition the trajectory into a set of $k$ segments indexed by $\tau_0=0, \tau_1, \ldots, \tau_k=T$. We define the value of the belief state at the beginning of a segment similarly to Equation~\ref{eqn:ValueFunction}, but we now accumulate the cost over $\tau_{i+1} - \tau_i$ steps, and take the expected value of the belief state at the end of $\tau_{i+1} - \tau_i$ steps:
\begin{equation}
    V^{\pi}(x_{\tau_i},b_{\tau_i}) =
    \mathbb{E}_{z\sim b_t} \left[
        \sum_{t=\tau_i}^{\tau_{i+1}-1}
            l(x_t,u_t,z)
            + \mathbb{E}_{o_{\tau_{i+1}-1},x_{\tau_{i+1}}} \left[
                V^{\pi}(x_{\tau_{i+1}},b_{\tau_{i+1}})
            \right]
    \right].
\label{eqn:HierarchicalValue}
\end{equation}
The second-order expansion can be taken similarly to before, but now with respect to perturbations of each segment. Hierarchical dynamic programming can be further optimized by applying standard DDP recursions to each step of a segment. Our experiments use hierarchical PODDP with $k=3$.

\section{Results}\label{sec:Experiment}

Our experiments analyze the performance of PODDP in three environments, designed to test the ability of PODDP to plan under uncertainty about goal locations, environment dynamics, and other agents' intentions, respectively. We compare PODDP against two baselines. The first baseline, ``Maximum-likelihood DDP'' (MLDDP), assumes the latent state with the highest probability is the true latent state, and runs standard DDP. At each observation point, it replans based on the updated most-likely belief. The second baseline, ``Probability weighted DDP'' (PWDDP), minimizes the expected cost of a control sequence with respect to the current belief -- this is straightforward to implement using a version of Equation~\ref{eqn:HierarchicalValue}, with $k=1$ and $\tau_k$ equal to the horizon length.

\subsection{Experiment 1: Planning under cost function uncertainty}\label{sec:Experiment_TMaze}

\begin{figure}[htb]
    \centering
    \includegraphics[width=\linewidth]{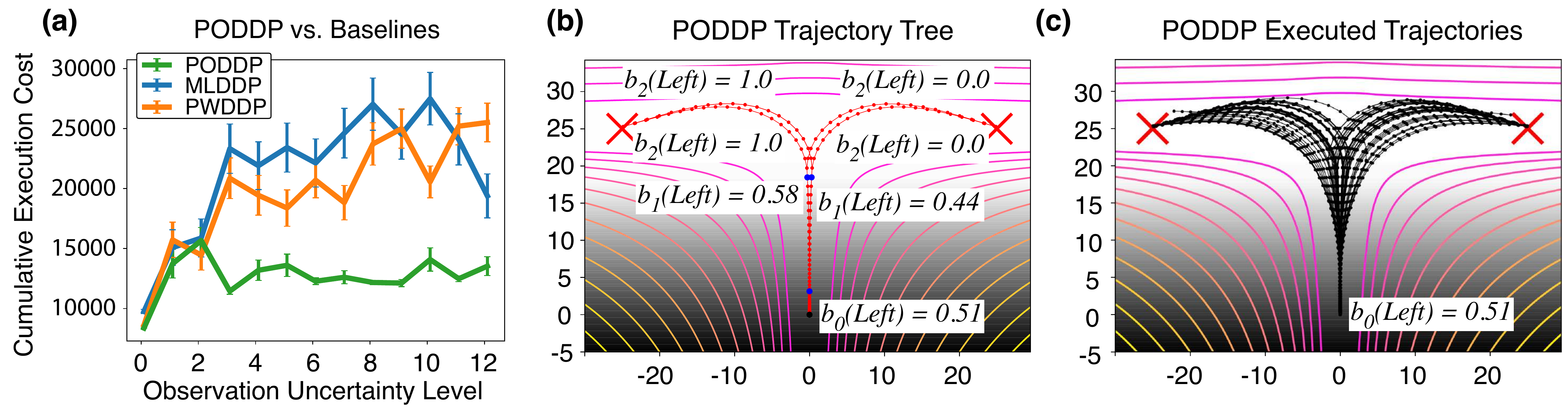}
    \caption{Experiment 1 results. (a) Average cumulative cost and variance of PODDP are lower than those of MLDDP and PWDDP across thirteen different observation uncertainty levels. Error bars show standard error. (b) Visualization of the PODDP trajectory tree in the T-Maze environment. (c) 100 sampled PODDP executions, used to compute the datapoint in (a) for observation uncertainty level = 9.1. See Supplementary for comparable executions of MLDDP and PWDDP.}
    \label{fig:Expt1}
\end{figure}

Our first experiment tests PODDP in a scenario in which the location of a goal is unknown, and determined by the latent world state. The environment is structured as a ``T-Maze'': a long corridor (surrounded by high cost regions), which splits left and right at the end. A binary latent state determines whether the goal is on the $ \textit{Left} $ or $ \textit{Right} $. Goal costs which increase quadratically with the distance from the true goal location induce the agent to move to the goal as quickly as possible. Fig.~\ref{fig:Expt1}(b) shows the environment, with a contour plot of the location cost overlaid, and goal locations marked with X's. The agent is a simulated vehicle with non-holonomic bicycle dynamics. The observation function generates a Gaussian random variable conditioned on the latent state $z$: the mean is $-1$ if $z=\textit{Left}$ and $1$ if $z=\textit{Right}$. The uncertainty of the observation decreases as the vehicle moves to the end of the maze; this uncertainty is parameterized by a smooth function which outputs the variance of the distributions, illustrated by the background gradient in Fig.~\ref{fig:Expt1}(b).

Fig.~\ref{fig:Expt1}(a) compares the average cumulative cost incurred by PODDP versus two baseline models over 100 sampled executions in each of thirteen environments, each with a different level of observation uncertainty. PODDP outperforms both baselines, and has lower variance.

Fig.~\ref{fig:Expt1}(b) shows a trajectory tree optimized by PODDP, starting from the belief $b(z = \textit{Left}) = 0.51$. The tree contains a contingency plan for all possible maximum-likelihood outcome sequences, conditioned on the latent state values. Fig.~\ref{fig:Expt1}(c) shows the 100 executed trajectories used in Fig.~\ref{fig:Expt1}(a) for uncertainty level $= 9.1$, sampling observations and state transitions from their true distributions. Among the executed trajectories are some in which the agent first moves to one side, then crosses back to seek the goal on the other side. These correspond to ``bad'' observations which indicate the incorrect latent state; Fig.~\ref{fig:Expt1}(b) shows that PODDP plans for these contingencies, and Fig.~\ref{fig:Expt1}(c) shows that it handles them gracefully, by responding conservatively to noisy observations so that recovery is possible following later, better observations.

Table~\ref{tbl:Expt1} shows the results of a targeted analysis on the mean cumulative cost incurred by each model over $1000$ executions for observation uncertainty level $= 9.1$. PODDP incurred significantly less mean cumulative cost than MLDDP ($t(1998) = 15.1, p < 0.00001$, and PODDP also incurred significantly less mean cumulative cost than PWDDP ($t(1998) = 18.9, p < 0.00001$). The mean cumulative costs incurred by MLDDP and PWDDP were not significantly different ($t(1998) = 0.05, p = 0.96$).

\begin{table}[h]
\centering
\caption{Mean cumulative cost (standard error in parentheses) incurred by each model in Experiment 1 over $1000$ samples.}
\begin{tabular}{ccc}
\hline
PODDP           & MLDDP           &  PWDDP                 \\
\hline
13330.6 (244.5) & 23839.8 (649.1) & 23878.5 (500.7) \\
\hline
\end{tabular}
\label{tbl:Expt1}
\end{table}

\subsection{Experiment 2: Planning under dynamical mode uncertainty}\label{sec:Experiment_MuddyTrack}

\begin{figure}[b]
    \centering
    \includegraphics[width=.7\linewidth]{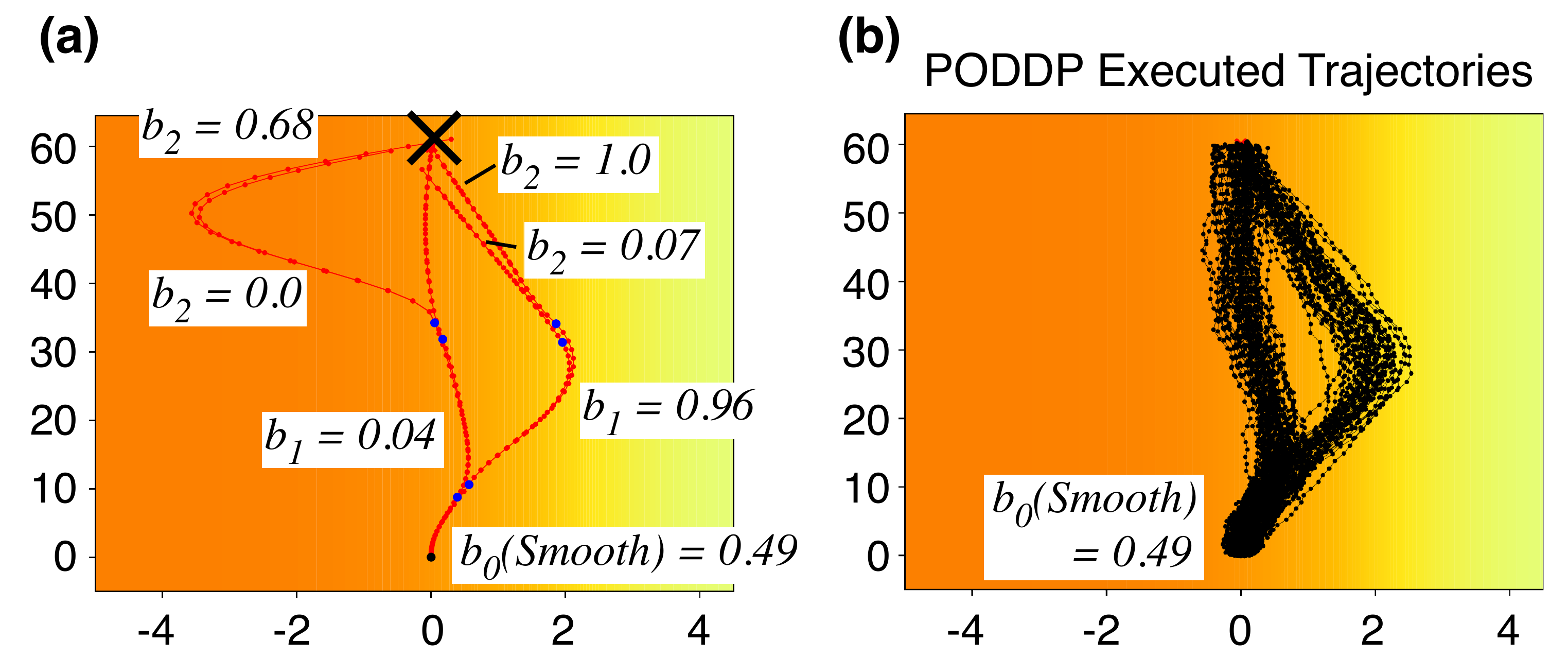}
    \caption{Experiment 2 results. (a) Visualization of the PODDP trajectory tree in the Rough Terrain environment. (b) 100 sampled PODDP executions. See Supplementary for comparable executions of MLDDP and PWDDP.}
    \label{fig:Expt2}
\end{figure}

Our second experiment tests whether PODDP can plan in the belief space over uncertain, partially observable dynamical modes of the environment. In this scenario, shown in Fig.~\ref{fig:Expt2}(a), a vehicle with non-holonomic bicycle dynamics is moving toward a goal (marked by an X) over rough terrain (e.g., ``mud''), which exerts a resistive force while the vehicle is moving, imposing cost due to the additional force required to maintain a constant velocity. A binary latent state determines the smoothness of the terrain to the right of the vehicle: When the latent state $z=\textit{Smooth}$, the terrain to the right exerts low resistive force; when $z=\textit{Rough}$, the terrain to the right is rough, with high resistive force equal to that on the left. Fig.~\ref{fig:Expt2}(a) shows the gradient from rough to smooth terrain going from left to right when the latent state is $\textit{Smooth}$.

The only source of information about the latent state comes from observing the dynamics themselves via the state sequence. This presents a challenging planning problem: exploring the environment to infer the value of $z$ requires a costly detour right into the potentially smooth area, but the payoff is large if the agent can learn that the terrain is smooth and reduce cost thereafter.

Fig.~\ref{fig:Expt2}(a) shows that PODDP plans an exploratory policy to learn the value of $z$. The planned trajectory, starting from the belief $b(z=\textit{Smooth}) = 0.49$, immediately moves to the right to gain information about $z$; the first observation yields strong information about $z$, and the beliefs become near certain, which the conditional plan can then exploit either by veering into the smooth area, or by heading directly through the mud to the goal location. Fig.~\ref{fig:Expt2}(b) shows 100 sampled executions through the rough terrain environment, demonstrating the robustness of the planned PODDP trajectory tree.

Table~\ref{tbl:Expt2} reports the mean cumulative cost over $1000$ executions for each model. PODDP incurred significantly lower mean cumulative cost than MLDDP, $(t(1998) = 3.9, p=0.00008)$, and PODDP also incurred significantly less mean cumulative cost than PWDDP $(t(1998) = 2.7, p=0.007)$. The mean cumulative costs incurred by MLDDP and PWDDP were not significantly different $(t(1998) = 0.41, p=0.68)$.

\begin{table}[h]
\centering
\caption{Mean cumulative cost (standard error in parentheses) incurred by each model in Experiment 2 over $1000$ samples.}
\begin{tabular}{ccc}
\hline
PODDP           & MLDDP           &  PWDDP                 \\
\hline
22110.7 (124.4) & 22710.7 (87.3) & 22639.5 (151.9) \\
\hline
\end{tabular}
\label{tbl:Expt2}
\end{table}

\subsection{Experiment 3: Latent intention-aware interactive lane changing} \label{sec:Experiment_LaneChange}

This experiment tested the ability of PODDP to plan trajectories through a belief state which includes the latent intentions of other agents, and dynamics which capture agents' intention-dependent actions. This scenario adds another vehicle to the state space, parameterized by a longitude and velocity (the planner vehicle again has bicycle dynamics). The other vehicle dynamics are modeled using a modified Intelligent Driver model (IDM), with a smooth boundary function for identifying the leading vehicle.
The latent state represents whether the other driver is $Nice$ or $Aggressive$. If the other driver is $Nice$, it is assumed to have a lower desired speed, and to slow down for others. If the other driver is $Aggressive$, it is assumed to have a higher desired speed, and to not slow down for others.

\begin{figure}[htb]
    \centering
    \includegraphics[width=.85\linewidth]{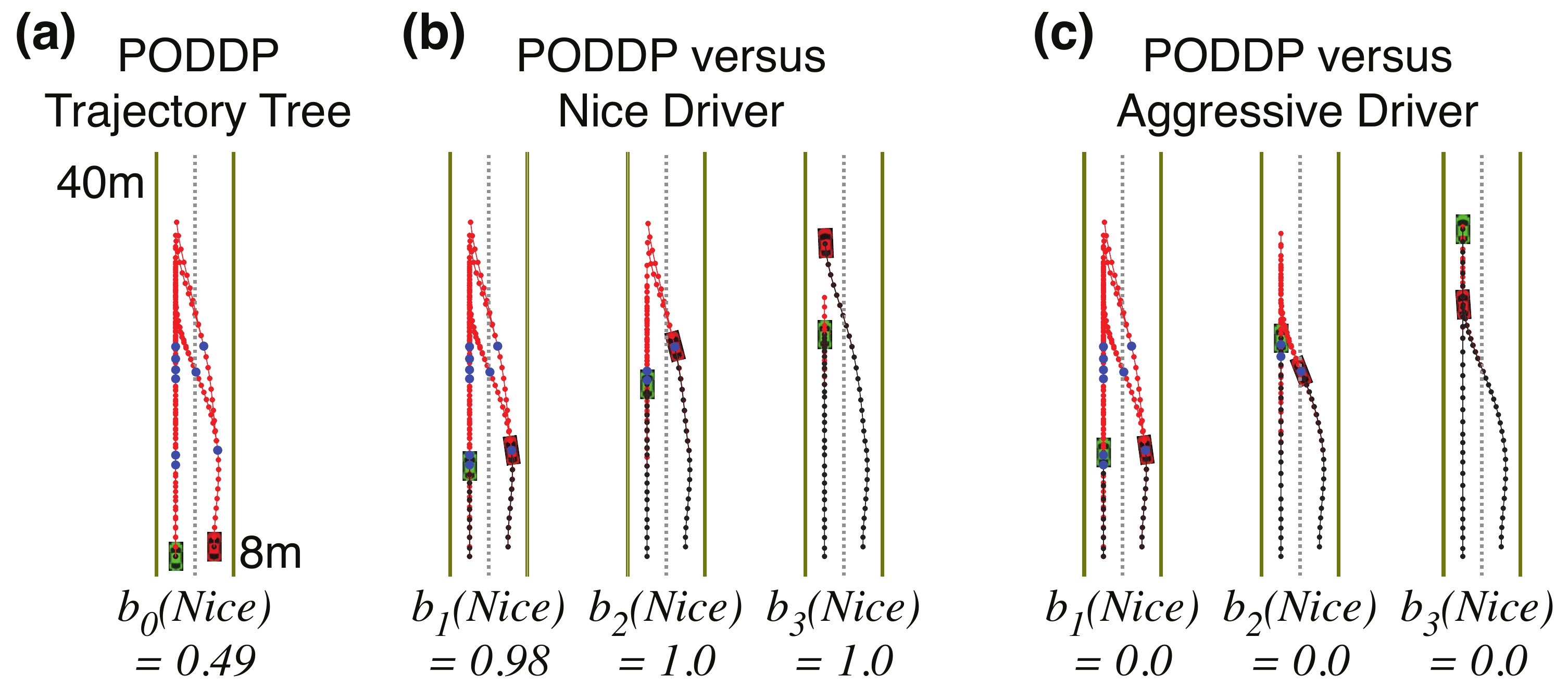}
    \caption{Experiment 3 results. (a) Visualization of the PODDP trajectory tree in the Lane Change environment. (b) Sampled PODDP execution against a $\textit{Nice}$ driver. PODDP successfully merges ahead of the other vehicle. (c) Sampled PODDP execution against an $\textit{Aggressive}$ driver. PODDP successfully merges behind the other vehicle. See Supplementary for videos of executed trajectories for PODDP, MLDDP and PWDDP.}
    \label{fig:Expt3}
\end{figure}

Fig.~\ref{fig:Expt3} shows that PODDP can plan in the belief space over the other vehicle's latent state, and can construct a contingency plan to change lanes ahead of the other vehicle if it is inferred to be $Nice$, or change lanes behind the other vehicle if it is inferred to be $Aggressive$. Fig.~\ref{fig:Expt3}(b) and (c) show the successful execution (black past trajectories) of these plans (red future trajectories). PWDDP also succeeds at changing lanes ahead of the $Nice$ driver, and changing lanes behind the $Aggressive$ driver. However, as shown in Table~\ref{tbl:Expt3}, over $1000$ sample executions, PWDDP incurred significantly higher cost than both PODDP $t(1998) = 14.8, p < 0.00001 $ and MLDDP $t(1998) = 8.3, p < 0.00001 $. In contrast, MLDDP fails to pass the $Nice$ driver, and always changes lanes behind both $Nice$ and $Aggressive$ drivers, and incurs significantly higher cost than PODDP $(t(1998) = 5.3, p < 0.00001)$.  This is because the maximum likelihood initial belief is $Aggressive$, which leads MLDDP to immediately decelerate, losing the chance to pass. To provide a fair comparison, we reran $1000$ sample executions with $b_0(Nice) = 0.51$. With this prior, MLDDP succeeds at passing the $Nice$ driver, and changing lanes behind the $Aggressive$ driver, but incurs a higher mean cumulative cost, as shown in Table~\ref{tbl:Expt3}. We also ran PODDP and PWDDP in this condition; as expected, the mean cumulative costs incurred were not significantly different than with the other prior.

\begin{table}[ht]
\centering
\caption{Mean cumulative cost (standard error in parentheses) incurred by each model in Experiment 3 over $1000$ samples.}
\begin{tabular}{cccc}
\hline
PODDP           & MLDDP           &  PWDDP    & MLDDP ($b_0(Nice) = 0.51$)            \\
\hline
121.3 (0.46) & 130.3 (1.6) & 152.1 (2.0)      & 143.2 (2.0) \\
\hline
\end{tabular}
\label{tbl:Expt3}
\end{table}

\section{Conclusion}

\label{sec:Conclusion}

We presented the PODDP algorithm for planning in POMDPs with continuous states, actions, and observations, nonlinear dynamics, and partially observable discrete latent states. PODDP is practical for many classes of problems: We demonstrated that it can perform belief-space planning in tasks with uncertainty about the (1) cost function, (2) dynamics, or (3) latent intentions of other agents. Future research will extend our work to handle partially observable continuous as well as discrete states, and also dynamic latent states, and further explore the range of applications these techniques enable.

\bibliographystyle{unsrt}
\bibliography{refs}

\newpage

\begin{appendices}

\maketitle

\section{Supplementary Experimental Methods and Results}

\subsection{Experiment 1}

\subsubsection{Model evaluation}

To quantitatively evaluate the model, we ran 100 executions in thirteen different environments, each with a different level of observation uncertainty in figure \ref{fig:Expt1}(a). The observation uncertainty levels used were $[0.1, 1.1, 2.1, \ldots, 12.1]$. Each execution sampled the ground truth world state from the prior (set to $P_0(\textit{Left}) = 0.49$), and sampled observations from the observation distribution at each observation timestep. All algorithms replanned after each observation. For this experiment, the vehicle dynamics were assumed to be deterministic, because they were independent of the latent state value, and not relevant for the task.

\subsubsection{Supplementary results}

\begin{figure}[htb]
    \centering
    \includegraphics[width=\linewidth]{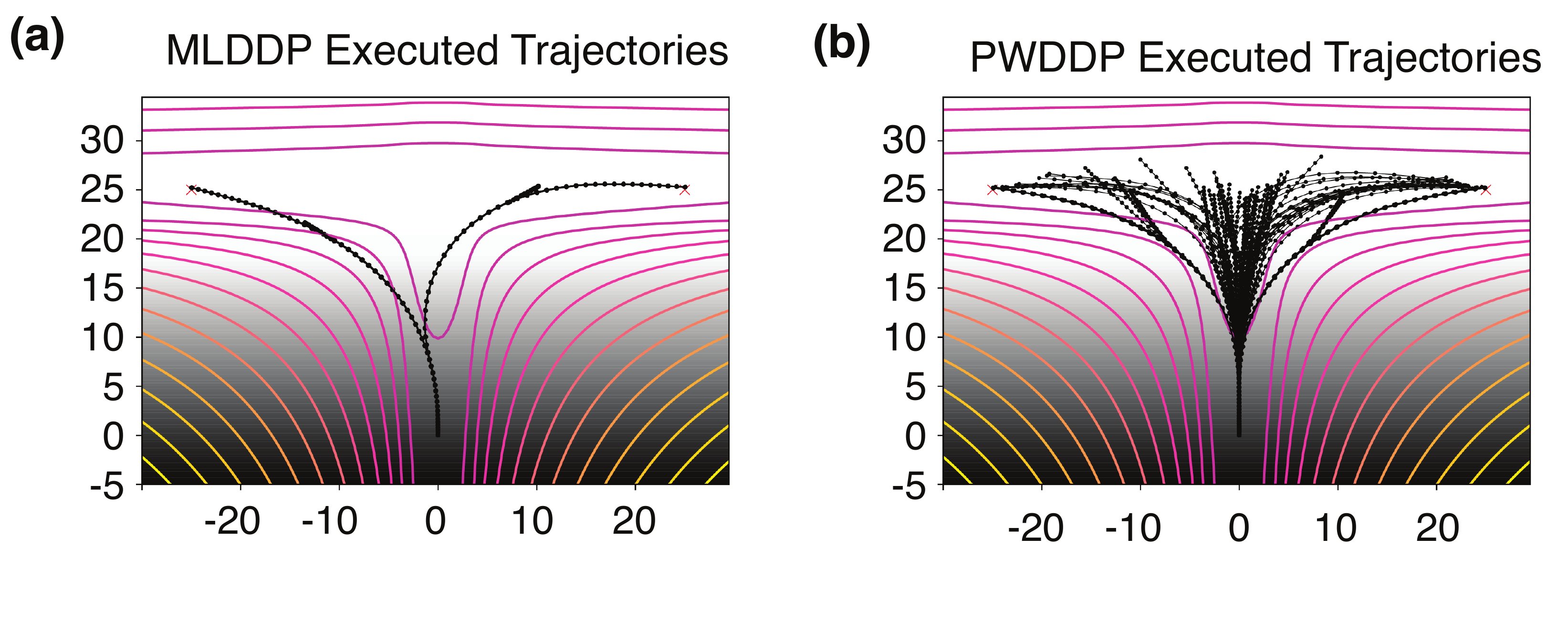}
    \caption{Experiment 1 results. (a) 100 sampled MLDDP executions. (b) 100 sampled PWDDP executions. Compare to Main Text, Fig.~\ref{fig:Expt1}(c).}
    \label{fig:SuppExpt1}
\end{figure}

Fig.~\ref{fig:SuppExpt1}(a) and (b) show sampled trajectories from MLDDP and PWDDP, respectively. Both algorithms fail to explore as aggressively as PODDP, which accelerates more rapidly in the beginning of the trajectory to get as reliable an observation as possible. The possible MLDDP trajectories split at each observation point, depending on the maximum-likelihood belief state following the observation. These trajectories commit strongly to the maximum-likelihood goal location, and when a bad observation occurs, they are unable to recover. The PWDDP trajectories also commit early, based on minimizing the goal costs to both locations simultaneously, weighted by their probability. Only when the initial observation is very strong can PWDDP can commit strongly.

\subsection{Experiment 2}

\subsubsection{Scenario formulation}

To capture the rough terrain in this experiment, we assume that the terrain exerts a location dependent resistive force on the vehicle. We model this force as a velocity-dependent deceleration $r$, such that:
\begin{equation*}
r = \rho * \tanh(v),
\end{equation*}
which the vehicle must overcome with its own acceleration, incurring cost.
The parameter $\rho$ varies according to the location. In our scenario, $\rho$ is high for rough terrain, and zero for smooth terrain. The transition in our environment is sigmoidal in the $x$-dimension.

In this scenario, because all information about the latent state comes from the state transitions, we assume additive Gaussian noise at each timestep for all state variables to make the belief state dynamics nontrivial.

\subsubsection{Supplementary results}

\begin{figure}[htb]
    \centering
    \includegraphics[width=\linewidth]{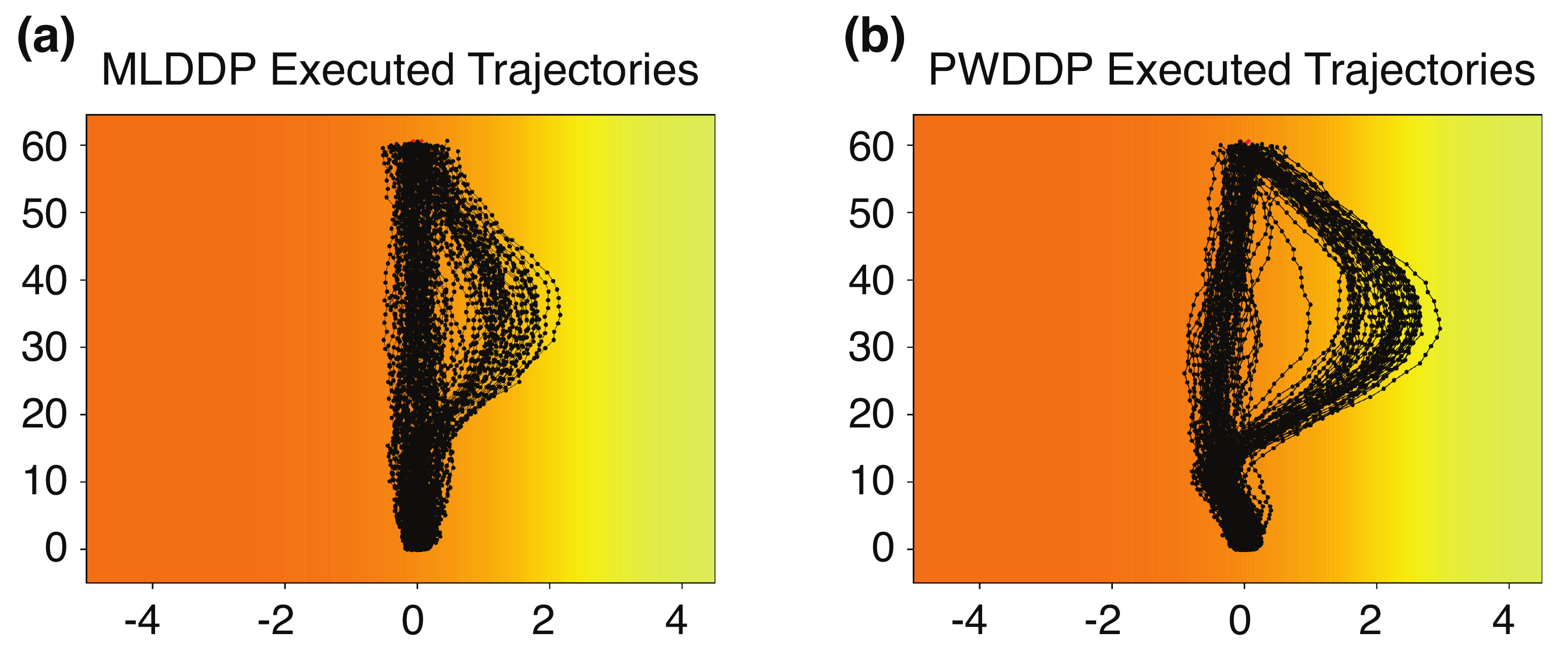}
    \caption{Experiment 2 results. (a) 100 sampled MLDDP executions. (b) 100 sampled PWDDP executions. Compare to Main Text, Fig.~\ref{fig:Expt2}(b).}
    \label{fig:SuppExpt2}
\end{figure}

Fig.~\ref{fig:SuppExpt2}(a) and (b) show sampled trajectories from MLDDP and PWDDP, respectively. Both algorithms fail to explore as aggressively as PODDP, which veers left into the potentially smooth area to obtain as reliable an observation as possible. MLDDP initially assumes that $z=\textit{Rough}$, and heads directly for the goal location. If the maximum-likelihood belief after the first observation is $z=\textit{Smooth}$, it moves into the smooth region. PWDDP actually moves away from the smooth region, because this trajectory allows it to minimize distance from the goal with the same control sequence under both dynamics. This is an interesting and subtle feature of PWDDP. It also moves into the smooth region if the belief updates in favor of $z=\textit{Smooth}$, but it does not explore or plan for future observation contingencies.

\subsection{Experiment 3}

\subsubsection{Scenario formulation}

In this scenario, we assume that both vehicles start from a velocity of 10 m/s. The dynamics of the other vehicle follow a modified IDM model.

In this scenario, because all information about the latent state comes from the other vehicle's state transitions, we assume additive Gaussian noise at each timestep for all state variables to make the belief state dynamics nontrivial.

\subsubsection{Supplementary results}

Please see \texttt{https://davidqiu1993.github.io/poddp-paper} for supplementary videos.

\todo{may include analysis or description of the results, and the comparison table}

\section{Derivation of Hessians from the backward pass}

PODDP computes a second-order approximation $\tilde{Q}$ to the $Q$-function defined in the main text, by taking first- and second-derivatives of the dynamics and cost functions with respect to perturbations of the belief state $\delta s$ and controls $\delta u$. For reasons of space and brevity, we derive the Hessian matrices here.

Note that following the standard iLQR approach~\cite{tassa2012}, we discard the terms involving Hessians of the dynamics $\frac{ \partial^2 s_{z}' }{ \partial \delta s^2 }$, $\frac{ \partial^2 s_{z}' }{ \partial \delta s \partial \delta u }$, and $\frac{ \partial^2 s_{z}' }{ \partial \delta u^2 }$.

The Hessians of the $\tilde{Q}$-function are:
\[
\begin{split}
Q_{ss} =
& \sum_{z \in \mathcal{Z}} \left[
    \frac{ \partial^{2} b_{z} }{ \partial \delta s^{2} } \left( l_{z} + V_{z}' \right) +
    2 \frac{ \partial b_{z} }{ \partial \delta s } \left(
        \frac{ \partial l_{z} }{ \partial \delta s} +
        \frac{ \partial s_{z}' }{ \partial \delta s }^{\mathsf{T}} \frac{ \partial V_{z}' }{ \partial s_{z}' }
    \right)^{\mathsf{T}} + \right. \\
&   \left. b_{z} \left(
        \frac{ \partial^{2} l_{z} }{ \partial \delta s^{2} } +
        \frac{ \partial s_{z}' }{ \partial \delta s }^{\mathsf{T}} \frac{ \partial^{2} V_{z}' }{ \partial s_{z}'^{2} } \frac{ \partial s_{z}' }{ \partial \delta s } +
        \frac{ \partial V_{z}' }{ \partial s_{z}' }^{\mathsf{T}} \frac{ \partial^{2} s_{z}' }{ \partial \delta s^{2} } \right)
\right]
\end{split}
\]
\[
\begin{split}
Q_{su} =
& \sum_{z \in \mathcal{Z}} \left[
    \frac{ \partial b_{z} }{ \partial \delta s } \left(
        \frac{ \partial l_{z} }{ \partial \delta u } +
        \frac{ \partial s_{z}' }{ \partial \delta u }^{\mathsf{T}} \frac{ \partial V_{z}' }{ \partial s_{z}' }
    \right)^{\mathsf{T}} + \right. \\
&   \left. b_{z} \left(
        \frac{ \partial^{2} l_{z} }{ \partial \delta s \partial \delta u } +
        \frac{ \partial s_{z}' }{ \partial \delta s }^{\mathsf{T}} \frac{ \partial^{2} V_{z}' }{ \partial s_{z}'^{2} } \frac{ \partial s_{z}' }{ \partial \delta u } +
        \frac{ \partial V_{z}' }{ \partial s_{z}' }^{\mathsf{T}} \frac{ \partial^{2} s_{z}' }{ \partial \delta s \partial \delta u }
    \right)
\right]
\end{split}
\]
\[
Q_{uu} = \sum_{z \in \mathcal{Z}} \left[
    b_{z} \left(
        \frac{ \partial^{2} l_{z} }{ \partial \delta u^{2} } +
        \frac{ \partial s_{z}' }{ \partial \delta u }^{\mathsf{T}} \frac{ \partial^{2} V_{z}' }{ \partial s_{z}'^{2} } \frac{ \partial s_{z}' }{ \partial \delta u } +
        \frac{ \partial V_{z}' }{ \partial s_{z}' }^{\mathsf{T}} \frac{ \partial^{2} s_{z}' }{ \partial \delta u^{2} }
    \right)
\right]
\]

\section{Derivation of value function recursion from the backward pass}

We can plug $ \delta u^{*} $ from Equation~\ref{eqn:ControlUpdate} in the main text back into $ \tilde{Q} $, to calculate the approximate quadratic model of the value function $ V $:
\begin{equation}
\begin{matrix}
    \Delta V \approx - \frac{1}{2} k^{\mathsf{T}} Q_{uu} k &
    \frac{ \partial V }{ \partial s } \approx Q_{s} - \frac{1}{2} K^{\mathsf{T}} Q_{uu} k &
    \frac{ \partial^{2} V }{ \partial s^{2} } \approx Q_{ss} - \frac{1}{2} K^{\mathsf{T}} Q_{uu} K .
\end{matrix}
\end{equation}
In Algorithm~\ref{alg:BackwardPass}, the $ \textsc{OptimizeControl} $ function returns $\Delta = \left\langle V, \frac{ \partial V }{ \partial s }, \frac{ \partial^{2} V }{ \partial \delta s^{2} } \right\rangle $.

\end{appendices}

\end{document}